\documentclass{article}

\usepackage[final]{neurips_2019}

\usepackage[utf8]{inputenc}
\usepackage[T1]{fontenc}
\usepackage{hyperref}
\usepackage{url}
\usepackage{float}
\usepackage{booktabs}
\usepackage{amsfonts}
\usepackage{nicefrac}
\usepackage{microtype}
\usepackage{graphicx}
\usepackage{xcolor}
\usepackage{lipsum}
\usepackage{enumitem}
\usepackage{amsmath}
\usepackage{caption}
\usepackage{mathtools}
\usepackage{indentfirst}
\usepackage{subcaption}

\title{
  2-Tier SimCSE: Elevating BERT for Robust Sentence Embeddings\\
}

\author{
  Aubrey (Yumeng) Wang\hspace{0.5cm} Candice (Junjin) Wang\hspace{0.5cm} Ziran Zhou\\
  Department of Management Science \& Engineering \\
  Stanford University \\
  \texttt{yumeng03@stanford.edu} \\ \texttt{candicew@stanford.edu}\\ \texttt{ziran@stanford.edu}\\
  Mentor: Arvind Mahankali
}

\begin{document}

\maketitle

\begin{abstract}
Effective sentence embeddings that capture semantic nuances and generalize well across diverse contexts are crucial for natural language processing tasks. We address this challenge by applying SimCSE (Simple Contrastive Learning of Sentence Embeddings) using contrastive learning to fine-tune the minBERT model for sentiment analysis, semantic textual similarity (STS), and paraphrase detection. Our contributions include experimenting with three different dropout techniques, namely standard dropout, curriculum dropout, and adaptive dropout, to tackle overfitting, proposing a novel 2-Tier SimCSE Fine-tuning Model that combines both unsupervised and supervised SimCSE on STS task, and exploring transfer learning potential for Paraphrase and SST tasks. Our findings demonstrate the effectiveness of SimCSE, with the 2-Tier model achieving superior performance on the STS task, with an average test score of 0.742 across all three downstream tasks. The results of error analysis reveals challenges in handling complex sentiments and reliance on lexical overlap for paraphrase detection, highlighting areas for future research. The ablation study revealed that removing Adaptive Dropout in the Single-Task Unsupervised SimCSE Model led to improved performance on the STS task, indicating overfitting due to added parameters. Transfer learning from SimCSE models on Paraphrase and SST tasks did not enhance performance, suggesting limited transferability of knowledge from the STS task.

\end{abstract}

% {\color{red} This template does not contain the full instruction set for this assignment; please refer back to the milestone instructions PDF.}

\section{Introduction}
In the realm of natural language processing (NLP), the quest for effective sentence embeddings stands as a pivotal challenge, essential for a multitude of downstream tasks. The challenge lies in crafting embeddings that not only capture semantic nuances but also generalize well across diverse linguistic contexts. Current methods often rely on labeled data for supervised learning, constraining scalability and applicability across languages and domains. Additionally, the discrete nature of language presents a formidable obstacle to devising contrastive learning frameworks similar to those successful in computer vision. This is where the crux of our effort lies: trying a new approach that takes advantage of large amounts of unlabeled text data and overcomes the limitations of the supervisory approach. Our approach capitalizes on unsupervised learning techniques to forge superior sentence embeddings, circumventing the need for task-specific labeled datasets. Central to our methodology is a contrastive learning framework adept at handling both unsupervised and supervised settings, thereby ensuring versatility across tasks and languages. Drawing inspiration from recent advancements in contrastive learning for computer vision tasks \citet{he2020momentum} and \citet{zhai2021contrastive}, we aim to adapt and extend these methodologies to the realm of NLP. In continuous representation learning and language model pre-training, we seek to address the inherent anisotropy problem plaguing language representations through theoretical insights and empirical validation.

\section{Related Work}
Natural language processing (NLP) research has been driven by the need to overcome challenges such as representation degeneration and anisotropy to find universal sentence embeddings. Since the anisotropy problem was identified, various methodologies have emerged to address it. Gao et al. \citet{gao2019anisotropic} proposed a regularization method to mitigate representation degeneration in models trained with tied word embedding matrices. Contrastive learning aims to bring similar instances closer and push dissimilar ones farther apart. Building on the work of Gao et al. \citet{gao2019anisotropic}, contrastive objectives have shown effectiveness in alleviating the anisotropy problem. The SimCSE framework provides an efficient solution for constructing positive pairs through contrastive learning, leveraging standard dropout for data augmentation in the unsupervised approach and entailment pairs from nli\_for\_simcse datasets for the supervised approach. This simplicity and effectiveness make SimCSE applicable across various NLP tasks, especially in unsupervised scenarios. With the same goal for improving model generalization, recent advancements in dropout techniques such as adaptive dropout and curriculum dropout have shown promise in enhancing model generalization and performance across various tasks: Curriculum provides the model a chance to dynamically increase the dropout rates during training to combat overfitting \citet{Morerio2017dropout}, while adaptive dropout using a binary belief network to dynamically set neuron-specific dropout probabilities \citet{NIPS2013_7b5b23f4}. However, while SimCSE has shown promising results on STS tasks, its efficacy on smaller within-task datasets and tasks beyond STS remains an open question. Our work aligns with this research landscape by evaluating the optimization effects of SimCSE on the minBERT model across various single-task and simultaneous multiple-task scenarios. Consequently, our endeavor represents a significant advancement in the quest for universal sentence embeddings. By shedding light on the intricacies of model optimization and its implications for both research and practical applications in NLP, our project serves as a promising next step in advancing the field.
\section{Approach}
        \textbf{3.1 Baseline} \\
       Our baseline consists of Multitask and Single-task versions of minBERT, which integrate sentence tokenization, embedding combinations, and a multi-head self-attention mechanism across 12 transformer layers. The models use linear layers for attention and position-wise feed-forward networks with dropout for pooled BERT embeddings generation. Multitask minBERT is tailored for simultaneously handling 3 NLP tasks, featuring a dropout layer to reduce overfitting and 3 heads, one for each task, are applied to make predictions, where the SST and Paraphrase tasks use a linear classifier separately while the STS task use cosine similarity with sigmoid scaling as basis. Single-task minBERT shares minBERT's architecture but allows for task-specific operation, where the model for each downstream task is individually trained and fine-tuned. This approach enables focused task execution while maintaining minBERT's efficiency.\\ 
       \textbf{3.2 Main Approach} \\
       After implementing MinBERT with the provided skeleton code and project handout, we integrated AdamW \citet{kingma2014adam} and achieved a satisfactory dev accuracy for SST task. We then developed our Multitask and Single-task baselines,  where we used Binary Cross Entropy loss for SST and paraphrase tasks and MSE loss for STS task, all of which achieved reasonable dev scores (Both results are in Experiments section). Furthermore, since the three single-task models demonstrated improved performance compared to the multitask baseline, our following extensions are applied on the single-task baseline models.
       \\ 
        \hspace*{1em}\textbf{3.2.1 Dropout Methods Experiments} - our original idea \\
        To tackle the overfitting issue observed across all tasks, we applied 3 dropout strategies (standard, curriculum, adaptive) to address overfitting on the single-task models and observed the best-performing dropout strategies being adaptive and standard dropout. (Results in Experiments). \\
       \hspace*{2em} \textbf{3.2.1.1 Curriculum Dropout}: We adopt "standout" from \citet{Morerio2017dropout}, dynamically increasing dropout rates during training to combat overfitting. This approach, unlike standard dropout, suggests that co-adaptation of feature detectors and subsequent overfitting are less likely in early training stages, thereby enhancing the model's generalization capabilities. \\
       \hspace*{2em} \textbf{3.2.1.2 Adaptive Dropout}: We implemented the adaptive dropout method from \citet{NIPS2013_7b5b23f4} (using its layers.py), using a binary belief network to set neuron-specific dropout probabilities dynamically, unlike the fixed-rate of traditional dropout. This technique aims to enhance network generalization and performance by adjusting its structure based on the input. \\
       \hspace*{1em}\textbf{3.2.2 SimCSE} \\
       We then initiated the implementation of the Unsupervised SimCSE and Supervised SimCSE extension using a contrastive learning framework to fine-tune the MinBERT model. (Details in Experiments)\\
       \hspace*{2em}\textbf{3.2.2.1 Unsupervised SimCSE}, which inputs the same sentence twice into a pre-trained language model, each time applying different independently sampled dropout masks which acts as minimal data augmentation, to generate two distinct embeddings and these embeddings are then used as positive pairs and other sentences in the same mini-batch as “negatives” to predict the same sentence in a contrastive learning setup.\\
       \hspace*{2em}\textbf{3.2.2.2 Supervised SimCSE}, which leverages labeled datasets (incorporates annotated pairs from natural language inference (NLI) datasets), using entailment pairs as positives and contradiction pairs as hard negatives within the contrastive learning framework.
       % \begin{itemize} [topsep=0pt,partopsep=0pt,parsep=0pt,itemsep=0pt]
       % \item The contrastive framework takes a cross-entropy objective with in-batch negatives*:
       % \end{itemize}
       % \begin{gather*}
       % \displaystyle \ell_{i} = -\log \frac{e^{\text{sim}(h_{i}, h_{i}^{+})/\tau}}{\sum_{j=1}^{N} e^{\text{sim}(h_{i}, h_{j}^{+})/\tau}}
       % \end{gather*}
       \begin{itemize}[nosep] % 'nosep' attempts to minimize spacing around the list itself
       \item The contrastive framework takes a cross-entropy objective with in-batch negatives*:
       \end{itemize}
       % Remove space before the equation
       {\setlength{\abovedisplayskip}{-5pt} % Adjusts space above the equation
       \setlength{\abovedisplayshortskip}{-5pt} % Adjusts space above the equation in "short" scenarios
       \setlength{\belowdisplayskip}{-5pt} % Adjusts space below the equation
       \setlength{\belowdisplayshortskip}{-5pt} % Adjusts space below the equation in "short" scenarios
       \begin{equation*}
            \ell_{i} = -\log \frac{e^{\text{sim}(h_{i}, h_{i}^{+})/\tau}}{\sum_{j=1}^{N} e^{\text{sim}(h_{i}, h_{j}^{+})/\tau}}
        \end{equation*}
        }
        \\
       \\We developed two separate functions to implement the unsupervised and supervised SimCSE algorithms. Specifically, taking into consideration that SimCSE was proved to be most successful with STS tasks in ~\citep{gao2021simcse}, we pre-trained a single-task minBERT model for Unsupervised SimCSE, then fine-tuned the Unsupervised SimCSE on the STS dataset, experimented it with both standard and adaptive dropout, and observed a best-performing Unsupervised SimCSE with standard dropout with a dev Pearson Correlation score of 0.716 on STS task. Meanwhile, we pre-trained a single-task minBERT model for Supervised SimCSE, then fine-tuned the Supervised SimCSE on the nli\_for\_simcse dataset, experimented it with both standard and adaptive dropout, and observed a best-performing Supervised SimCSE with standard dropout with a dev Pearson Correlation score of 0.806 on STS task. \\
       \hspace*{1em}\textbf{3.2.3 Transfer Learning} - our original idea \\
       We then loaded the model weights of the best-performing Unsupervised SimCSE with standard dropout on STS task as pre-trained model, and fine-tuned it on the Paraphrase task and SST task separately, and achieved a dev accuracy of 0.777 and 0.526 respectively. Meanwhile, We loaded the model weights of the best-performing Supervised SimCSE with default dropout on STS task as pre-trained model, and fine-tuned it on the Paraphrase task and SST task separately, and achieved a dev accuracy of 0.782 and 0.487 respectively. \\
       \hspace*{1em}\textbf{3.2.4 2-Tier SimCSE Fine-tuning Procedure} - our original idea \\
       We began by pre-training a minBERT model for the STS task using the STS dataset. Following this initial step, we applied an unsupervised SimCSE technique for further fine-tuning on the same dataset. Subsequently, this enhanced model served as a foundation for a supervised SimCSE implementation, which underwent additional fine-tuning with the nli\_for\_simcse dataset to assess its STS task performance. Having established this 2-Tier SimCSE Fine-tuning Model, we then extended its application to both the Paraphrase and the SST tasks, employing transfer learning techniques to adapt the model effectively. Finally, We achieved a dev Pearson score of 0.811 on STS task, a dev accuracy of 0.785 on Paraphrase task, and a dev accuracy of 0.490 on SST task. (As shown in Figure 1 for detailed architecture for this 2-Tier SimCSE Fine-tuning Model)
       \begin{figure}[H]
           \centering
           \includegraphics[width=0.8\textwidth]{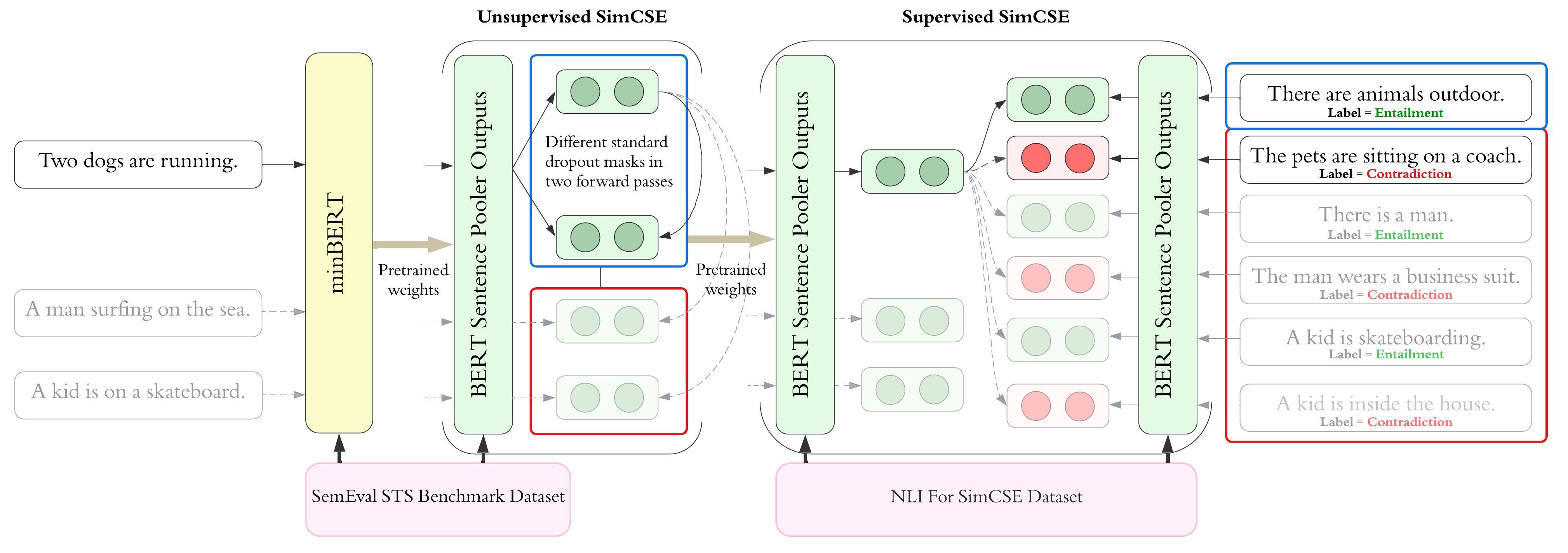}
           \caption{2-Tier SimCSE Fine-tuning Procedure}
           \label{fig:2-Tier SimCSE Fine-tuning Procedure}
       \end{figure}
       \begin{itemize} 
           \item{Note: We based our minBERT Baseline on provided skeleton code, adapted \textit{layers.py} from \citet{NIPS2013_7b5b23f4}, and referenced \textit{models.py }from ~\citep{gao2021simcse}. All other code were original code developed by ourselves.}
       \end{itemize} 
    
\section{Experiments}
\textbf{4.1 Data}
For sentiment classification, we used Stanford Sentiment Treebank (SST) (and CFIMDB) datasets for Sentiment Analysis, Quora dataset for Paraphrase Detection, SemEval STS Benchmark datasets for Semantic Textual Similarity (STS), all of which have already been described in detail in the handout.\\
Besides, we applied NLI dataset in the implementation of supervised SimCSE for STS downstream task. NLI dataset consists of SNLI and MNLI datasets. The SNLI corpus contains 570k human-written English sentence pairs labeled for balanced classification with the labels entailment, contradiction, and neutral\citet{snli}. MNLI dataset is a crowd-sourced collection of 433k sentence pairs annotated with textual entailment information \citet{multinli}. Modeled after the SNLI corpus, the corpus covers a range of genres of spoken and written text and allows for cross-genre generalization evaluation. The general NLI dataset consists of 275,000 examples, where each example consists of a sentence, a positive example, and a negative example\citet{gao2021simcse}.\\
\\
\textbf{4.2 Evaluation method}
As described in the leaderboard for results, the evaluation metrics employed for downstream tasks are accuracy for SST task and paraphrase task, alongside the Pearson Correlation score for STS task. In addition, we consider the trade-off between these metrics and the additional number of parameters and training time when training multiple BERT instances simultaneously or assembling separate models.

\textbf{4.3 Experimental Details and Results}\\
    \hspace*{1em}\textbf{5.3.1 Baselines}
    Table 2 presents a performance comparison between the Multitask baseline and the Single-task baselines, both using the best-performing cosine similarity with sigmoid scaling as the basis for STS task (selected from the 5 similarity measurements shown in Table 1) and default training parameters (learning rate of 1e-5, dropout rate of 0.3, weight decay of 0.0, batch size of 8, epochs of 10). The results demonstrate that the Single-Task Baseline outperformed the Multi-Task Baseline. This was somewhat expected, which is likely due to their focused optimization on specific tasks. This specialization allows the Single-Task Baselines to better capture task-specific patterns, leading to improved performance compared to the Multi-Task Baseline, which divides its capacity across multiple tasks. This finding highlights the well-known trade-off between specialization and generalization in multi-task learning.

    \vspace{-5pt}
    \begin{table}[H]
    \centering
    \resizebox{\textwidth}{!}{%
    \begin{tabular}{lll}
    \hline
    \textbf{Method} &
      \textbf{Description} &
      \textbf{Result} \\ \hline
    Sum of two sentence pair embeddings with linear trasnformation & 
      \begin{tabular}[c]{@{}l@{}}--\end{tabular} &
      0.381 \\
    Cosine similarity with simple scaling multiplication & 
      \begin{tabular}[c]{@{}l@{}}((cosine similarity score + 1) * 2.5)\end{tabular} &
      0.480 \\ 
    Cosine similarity with sigmoid scaling & 
      \begin{tabular}[c]{@{}l@{}}(sigmoid(cosine similarity score) * 5)\end{tabular} &
      0.502 \\ 
    Scaled-up cosine similarity with sigmoid scaling & 
      \begin{tabular}[c]{@{}l@{}}(sigmoid(cosine similarity score * 5) * 5)\end{tabular} &
      0.477 \\
      Cross-attention between sentence pair embeddings & 
      \begin{tabular}[c]{@{}l@{}}--\end{tabular} &
      0.338 \\ \hline
    \end{tabular}%
    }
    \caption{Scaling Methods for Cosine Similarity}
    \label{tab:scaling-methods}
    \end{table}
    \vspace{-20pt}
    \begin{table}[H]
    \centering
    \begin{tabular}{@{}lllll@{}}
    \toprule
    Model                     & Paraphrase Acc. & Sentiment Acc. & STS Corr. &  \\ \midrule
    Single-Task Baseline & 0.794           & 0.534          & 0.502     &  \\
    Multi-Task Baseline    & 0.743           & 0.477          & 0.429   
    &  \\ \bottomrule
    \end{tabular}
    \caption{Single-Task Baseline v.s. Multi-Task Baseline}
    \end{table}
    \vspace{-20pt}
     \hspace*{1em}\textbf{4.3.2 Dropout Methods}
    To improve generalization, we applied three different dropout techniques (standard, curriculum, and adaptive) to regularize the better-performing single-task models. We used default training parameters for SST and STS tasks and we changed the number of epochs to 5 while keeping all other parameters at their default values for paraphrase task. We observed the best-performing dropout strategies being adaptive dropout (highest performance on STS task) and standard dropout (highest performance on paraphrase and SST tasks), as shown in Table 3. This was not as what we expected that there was not one dropout method outperforms others, for which we will elaborate more on what may cause this issue during the following Analysis part.
    
    \vspace{-5pt}
    \begin{table}[H]
    \centering
    \begin{tabular}{@{}lllll@{}}
    \toprule
    Model                     & Paraphrase Acc. & Sentiment Acc. & STS Corr. &  \\ \midrule
    Single-Task Baseline with Standard Dropout & 0.794           & 0.534          & 0.502     &  \\
    Single-Task Baseline with Adaptive Dropout   & 0.785           & 0.522          & 0.579   &  \\
    Single-Task Baseline with Curriculum Dropout   & 0.777           & 0.527          & 0.500   
    &  \\ \bottomrule
    \end{tabular}
    \caption{Single-Task Baseline with Standard v.s. Adaptive v.s. Curriculum Dropout}
    \end{table}
    \vspace{-20pt}
    \hspace*{1em}\textbf{4.3.3 Unsupervised SimCSE}
    We pre-trained a single-task minBERT model for Unsupervised SimCSE and then fine-tuned the Unsupervised SimCSE on the STS dataset. Both the pre-training and fine-tuning stages used a batch size of 64, a learning rate of 3e-5, and a dropout probability of 0.1, while keeping all other parameters at their default values. We experimented with both standard and adaptive dropout versions. The best-performing Unsupervised SimCSE model, which used standard dropout, achieved a dev Pearson score of 0.716 on the STS task, as shown in Table 4. This was also not what we expected, and we will elaborate more on what may cause this issue during the following Analysis part.
    \vspace{-5pt}
    \begin{table}[H]
    \centering
    \begin{tabular}{@{}lllll@{}}
    \toprule
    Model                     &  STS Corr. &  \\ \midrule
    Single-Task Unsupervised SimCSE with Standard Dropout & 0.716          &  \\
    Single-Task Unsupervised SimCSE with Adaptive Dropout    & 0.656   
    &  \\ \bottomrule
    \end{tabular}
    \caption{Single-Task Unsupervised SimCSE with Standard v.s. Adaptive Dropout on STS Task}
    \end{table}
    \vspace{-20pt}
    \hspace*{1em}\textbf{4.3.4 Supervised SimCSE}
    We pre-trained a single-task minBERT model for Supervised SimCSE and then fine-tuned the Supervised SimCSE on the nli\_for\_simcse dataset, which consists of 275,000 data instances. Each instance in the nli\_for\_simcse dataset includes a sentence, its positive example sentence, and its negative example sentence. For both the pre-training and fine-tuning stages, we used a batch size of 24, a learning rate of 5e-5, a dropout probability of 0.1, and trained for 5 epochs. We experimented with both standard and adaptive dropout versions. The best-performing Supervised SimCSE model, which used standard dropout, achieved a dev Pearson score of 0.806 on the STS task. Additionally, we attempted to add an extra fine-tuning layer on the STS dataset instead of directly evaluating the STS task performance after fine-tuning the Supervised SimCSE on the nli\_for\_simcse dataset. However, this approach resulted in lower performance, as shown in Table 5. This was not what we had anticipated, and in the Analysis section that follows, we will go into further detail about the possible causes of this problem.
    \vspace{-5pt}
    \begin{table}[H]
    \centering
    \begin{tabular}{@{}lllll@{}}
    \toprule
    Model                     &  STS Corr. &  \\ \midrule
    Single-Task Supervised SimCSE with Standard Dropout & 0.806          &  \\
    Single-Task Supervised SimCSE with Adaptive Dropout    & 0.782   
    &  \\ 
    Single-Task Supervised SimCSE with Standard Dropout \\with additional fine-tuning on STS dataset & 0.725          &  \\
    Single-Task Supervised SimCSE with Adaptive Dropout \\with additional fine-tuning on STS dataset   & 0.731   
    &  \\ \bottomrule
    \end{tabular}
    \caption{Single-Task Supervised SimCSE with additional fine-tuning on STS dataset with Standard v.s. Adaptive Dropout on STS Task}
    \end{table}
    \vspace{-20pt}
    \hspace*{1em}\textbf{4.3.5 Transfer Learning}
    We then loaded the model weights of the best-performing Unsupervised SimCSE with standard dropout on the STS task as a pre-trained model and fine-tuned it separately on the Paraphrase and SST tasks. This approach achieved dev accuracies of 0.777 and 0.526 for the Paraphrase and SST tasks, respectively. Similarly, we loaded the model weights of the best-performing Supervised SimCSE with standard dropout on the STS task as a pre-trained model and fine-tuned it separately on the Paraphrase and SST tasks. (For both transfer learning processes, we used default training parameters for SST task and we changed the number of epochs to 5 while keeping all other parameters at their default values for paraphrase task) This resulted in a dev accuracy of 0.782 for the Paraphrase task and 0.487 for the SST task, as shown in Table 6. The result was not as what we expected, and we will elaborate more on what may cause this issue during the following Analysis part.\\
    \vspace{-10pt}
    \begin{table}[H]
    \centering
    \begin{tabular}{@{}lllll@{}}
    \toprule
    Model                     & Paraphrase Acc. & SST Acc. &  STS Corr. &  \\ \midrule
    Single-Task Unsupervised SimCSE\\ with Standard Dropout & 0.777           & 0.526       & 0.716    &  \\
    Single-Task Supervised SimCSE\\ with Standard Dropout  & 0.782           & 0.487       & 0.806
    &  \\ \bottomrule
    \end{tabular}
    \caption{Single-Task Unsupervised v.s. Supervised SimCSE with Standard Dropout}
    \end{table}
    \vspace{-20pt}
    \hspace*{1em}\textbf{4.3.6 2-Tier SimCSE Fine-tuning Model}
    We began by pre-training a minBERT model for the STS task using the STS dataset using default training parameters. Then we applied an unsupervised SimCSE technique for further fine-tuning on the same dataset using the best-performing model parameters (a batch size of 64, a learning rate of 3e-5, and a dropout probability of 0.1, while keeping all other parameters at their default values). Subsequently, this enhanced model served as a foundation for a supervised SimCSE implementation, which underwent additional fine-tuning with the nli\_for\_simcse dataset using the best-performing model parameters (a batch size of 24, a learning rate of 5e-5, a dropout probability of 0.1, and trained for 5 epochs) to assess its STS task performance. Having established this 2-Tier SimCSE Fine-tuning Model, we then extended its application to both the Paraphrase and the SST tasks using default training paramters for SST task and changing the number of epochs to 5 while keeping all other parameters at their default values for paraphrase task, employing transfer learning techniques to adapt the model effectively. Finally, We achieved a dev Pearson score of 0.811 on STS task, as shown in Table 7, which is impressive and better than expected. This innovative approach demonstrates the robustness and adaptability of combining multiple contrastive learning strategies for enhancing sentence embedding quality and STS task performance. We also achieved a dev accuracy of 0.785 on Paraphrase task, and a dev accuracy of 0.490 on SST task, as shown in Table 7, which is worse that we expected, likely due to the SimCSE framework, particularly in its 2-Tier implementation, significantly boosts performance in tasks closely aligned with its training objectives (STS task), which might not be transferable to the Paraphrase and SST tasks. This outcome implies a need for further exploration on task-specific model adaptations specifically for the Paraphrase and SST tasks as well.
    % \captionsetup[table]{aboveskip=0pt,belowskip=0pt}
    % \vspace{-10pt}
    % \begin{table}[H]
    %     \centering
    %     \begin{tabular}{@{}lllll@{}}
    %     \toprule
    %     Model & Paraphrase Acc. & SST Acc. & STS Corr. & \\ \midrule
    %     Single-Task 2-Tier SimCSE Fine-tuning Model with Default Dropout & 0.785 & 0.490 & 0.811 & \\
    %     \bottomrule
    %     \end{tabular}
    %     \caption{Single-Task 2-Tier SimCSE Fine-tuning Model with Default Dropout}
    % \end{table}
    % \vspace{-10pt}
    \captionsetup[table]{aboveskip=10pt, belowskip=10pt} % Adjust space around the caption

    % Add space before the table if needed (optional)
    \vspace{-5pt} % Increase space before the table
    \begin{table}[H]
        \centering
        \begin{tabular}{@{}lllll@{}}
        \toprule
        Model & Paraphrase Acc. & SST Acc. & STS Corr. & \\ \midrule
        Single-Task 2-Tier SimCSE Fine-tuning Model\\ with Standard Dropout & 0.785 & 0.490 & 0.811 & \\
        \bottomrule
        \end{tabular}
        \caption{Single-Task 2-Tier SimCSE Fine-tuning Model with Standard Dropout}
    \end{table}
    \vspace{-20pt} % Increase space after the table

    \hspace*{1em}\textbf{4.3.7 Test Leaderboard Scores}
    Figure 2 presents an accuracy comparison of different models on the Paraphrase, Sentiment (SST), and Semantic Textual Similarity (STS) tasks.\\
    We achieved the following scores with our best performing single-task models, 2-Tier SimCSE Fine-tuning Model on STS Task and Single-Task Baseline with Standard Dropout on Paraphrase and SST tasks, on Test Leaderboard:
    \begin{itemize} [topsep=0pt,partopsep=0pt,parsep=0pt,itemsep=0pt]
        \item SST test accuracy: 0.533
        \item Paraphrase test accuracy: 0.798
        \item STS test correlation: 0.788
        \item Overall test score: 0.742
    \end{itemize}

\begin{figure}[H]
    \centering
    \begin{subfigure}{0.48\textwidth}
        \centering
        \includegraphics[width=\textwidth]{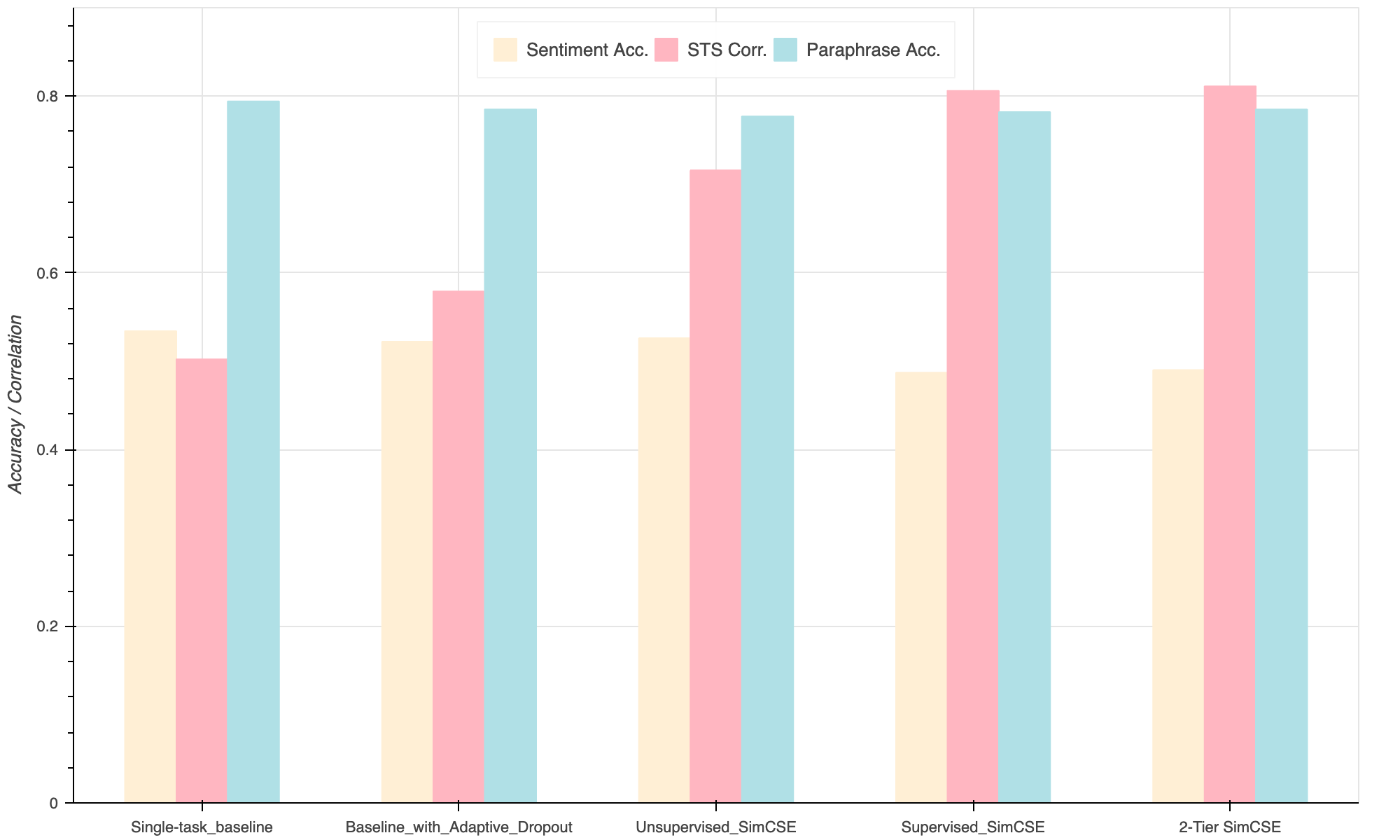}
        \caption{Figure 2: Accuracy comparison of different models on paraphrase sentiment, and STS tasks. }
    \end{subfigure}
    \hfill
    \begin{subfigure}{0.48\textwidth}
        \centering
        \includegraphics[width=\textwidth]{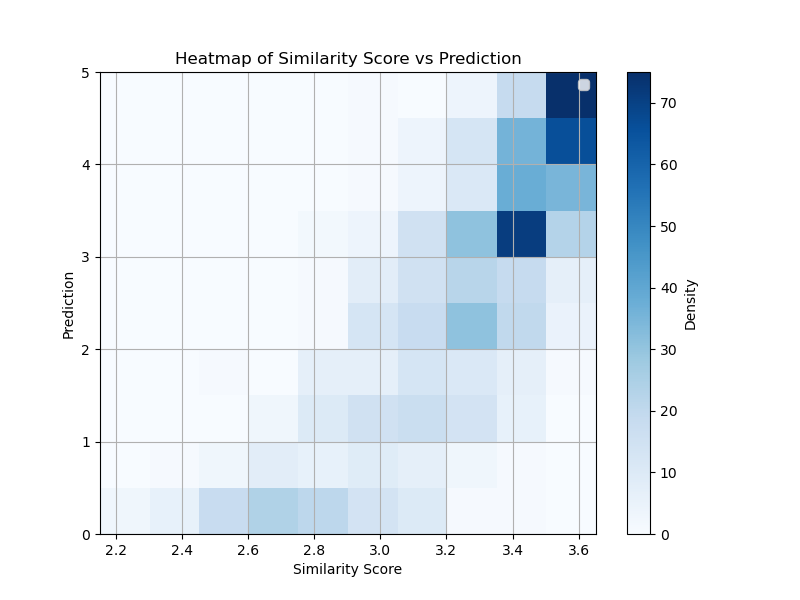}
        \caption{Figure 3: Heatmap of Similarity versus Prediction }
    \end{subfigure}
    \label{fig:both_images}
\end{figure}

\section{Analysis}
\textbf{5.1 Error Analysis}\\
\\
Upon examining the model outputs, we identified several common error patterns. For the SST task, there are primarily two error patterns. First, the models sometimes misclassified sentences with complex or mixed sentiments. For example, the sentence "On the heels of The Ring comes a similarly morose and humorless horror movie that, although flawed, is to be commended for its straight-ahead approach to creepiness." was classified as 0 (Negative) while the true label is 3 (Somewhat Positive), likely due to the presence of more negative words like "morose", "humorless" and "flawed," overshadowing the latter part of the sentence which carries a positive evaluation, including the positive phrase "to be commended." This highlights the challenge of accurately capturing nuanced sentiment, where the overall sentiment of a sentence can be subtly modulated by context, contrastive structures, and the cumulative effect of positive and negative expressions. Second, the true labels themselves can be ambiguous or questionable. For example, the sentence "And if you're not nearly moved to tears by a couple of scenes , you've got ice water in your veins." was classified as 4 (Positive) while the true label is 3 (Somewhat Positive), likely due to the expressive language used, such as "moved to tears" and "ice water in your veins," which is more of a conditional and metaphorical language expression rather than a direct wholly positive sentiment.\\
\\
As for the paraphrase detection task, many errors stem from the QQP dataset's borderline cases, such as questions with multiple parts or slight variations. For example, the model predicts "Why are Facebook, Google, and others not allowed in China?" as a paraphrase of "Why are Google, Facebook, YouTube and other social networking sites banned in China?" The model sometimes overfits to the word embeddings and relies too heavily on lexical overlap to determine paraphrases, without considering the overall semantic similarity. This is evident in the example where "What are some ways to register with Star Alliance?" is predicted as a paraphrase of "What can you get as a customer of Star Alliance?" Although both sentences are related to Star Alliance membership, the pre-trained BERT should have learned the difference to perform well on the Masked Language Modeling (MLM) task. Additionally, the model wrongly predicts unrelated sentences as paraphrases due to insufficient training data, as evident from the low occurrence of certain words in the dataset. For instance, the model incorrectly predicts "How does one copyright a program?" as a paraphrase of "Why does rocket reach maximum efficiency when vector velocity equals exhaust velocity?" The presence of synonym words for "equal" in the dataset, such as "fair," may also contribute to incorrect paraphrase predictions. Increasing the sample size of training data can help address the out-of-vocabulary issue and improve the model's performance on the paraphrase detection task.

For the STS task, the 2-Tier SimCSE Fine-tuning Procedure, which combines both unsupervised and supervised SimCSE techniques, generally performs well in capturing semantic similarity between sentence pairs. However, as Figure 3 (the heatmap of true similarity scores versus predictions for the 2-Tier SimCSE Fine-tuning Procedure) shows, the model is better at predicting sentence pairs with high (3-5) similarity scores than low (0-2) similarity scores, as evidenced by the relatively darker regions along the diagonal. This suggests that the model may struggle to make distinctions between moderately similar sentences or the training dataset is highly biased.
\\
\\
\textbf{5.2 Model Architecture Analysis}\\
\\
\hspace*{1em}\textbf{5.2.1 SimCSE}
SimCSE proves to be a powerful and efficient tool for generating enhanced universal sentence embeddings in our approach. The contrastive learning framework enables the model to learn representations that effectively capture syntactic and semantic information. The unsupervised SimCSE utilizes dropout as data augmentation, promoting the learning of robust and generalized representations from unlabeled data. The supervised SimCSE leverages labeled data to capture nuanced and task-specific semantic relationships. The success of SimCSE can be attributed to its ability to learn from large amounts of data, adapt to task-specific requirements, and generate high-quality sentence embeddings. Our 2-Tier SimCSE Fine-tuning Model, stacking both unsupervised and supervised techniques, achieves impressive performance on the STS task, surpassing the results of the single-task model fine-tuned specifically for STS task. This demonstrates the effectiveness of the SimCSE framework in capturing rich semantic information and generating high-quality sentence embeddings.
\\
\hspace*{1em}\textbf{5.2.2 Adaptive Dropout Ablation Study} Surprisingly, although Adaptive Dropout proved to have the highest performance on the STS task for the Single-Task Baseline Model among the three dropout strategies we experimented with, removing Adaptive Dropout in the subsequent Single-Task Unsupervised SimCSE Model on the STS task resulted in an increase in performance. We attribute this unexpected finding to the issue of overfitting, which we observed by noting a gap between the training and dev Pearson scores during the training process of Single-Task Unsupervised/Supervised SimCSE with Adaptive Dropout. In contrast, the training and dev Pearson scores matched well in the implementation of Single-Task Unsupervised/Supervised SimCSE with Standard Dropout. One possible reason for this is that Adaptive Dropout introduces four additional parameters: the previous layer, beta, previous state, and current state. These parameters are different for each initialization, leading to a significant increase in the total number of parameters in the model. The effect of the increased overfitting due to the substantial rise in the number of parameters outweighed the benefit of the dropout mechanisms in reducing overfitting, highlighting the importance of considering the trade-off between technique benefits and model complexity.
\\
\hspace*{1em}\textbf{5.2.3 Transfer Learning Ablation Study}
Comparing the performance of the Single-Task Baseline model with and without transfer learning from the SimCSE models on Paraphrase and SST Tasks showed that transfer learning did not provide a boost in performance on the Paraphrase and the SST task, but showed a decreased performance. This suggests that the knowledge gained from the STS task through SimCSE might not be transferable to the Paraphrase and SST task. While SimCSE learns universal sentence embeddings, transfer tasks sometimes require specialized or task-specific knowledge and understanding that goes beyond general semantics. Also, SimCSE builds on pre-trained minBERT. While minBERT has provided a strong foundation, its initial pre-training might not cover all aspects necessary for specific transfer tasks, therefore affecting the downstream performance of SimCSE.

\section{Conclusion}
We explored the application of SimCSE for fine-tuning minBERT on sentiment analysis, STS, and paraphrase detection tasks. We experimented with various dropout techniques, transfer learning potentials, and proposed a novel 2-Tier SimCSE Fine-tuning Model. Our main findings demonstrate the effectiveness of SimCSE in generating high-quality sentence embeddings, with the 2-Tier model achieving superior performance on the STS task. Future work could explore integrating advanced regularization techniques, applying SimCSE to other downstream tasks, and addressing the challenges revealed by error analysis, such as the models' difficulty in handling complex or mixed sentiments and their reliance on lexical overlap for paraphrase detection. We could focus on developing alternative architectures \citet{yin2019comparative} or attention mechanisms \citet{dasgupta2019dynamic} that may be better suited to capturing subtle differences in meanings or handling complex or mixed sentence structures, or incorporating additional features to address these limitations and improve the models' ability to capture semantic nuances and handle ambiguous or challenging cases. Overall, our project demonstrates the potential of contrastive learning and SimCSE in enhancing sentence embedding quality and downstream task performance. The proposed 2-Tier SimCSE Fine-tuning Model provides valuable insights for future research in this field.

\bibliographystyle{acl_natbib}
\bibliography{references}

\appendix

\section{Appendix}

\begin{table}[h]
\centering
\begin{tabular}{@{}lllll@{}}
\toprule
SST pretrain dev Acc. & SST finetune dev Acc. & CFIMDB pretrain dev Acc. & CFIMDB finetune dev Acc. &  \\ \midrule
0.409 & 0.521 & 0.788 & 0.976 &  \\ 
\bottomrule
\end{tabular}
\caption{Results on SST and CFIMDB Datasets for PART 1 of the project}
\end{table}
\end{document}